\documentclass[runningheads]{llncs}

\usepackage{eccv}

\usepackage{eccvabbrv}
\usepackage{multirow}
\usepackage{graphicx}
\usepackage{booktabs}

\usepackage[accsupp]{axessibility}  

\usepackage{hyperref}

\usepackage{orcidlink}

\begin{document}

\title{YesTrack: Referring Multi-Object Tracking via MLLM-based Yes/No Verification} 

\titlerunning{YesTrack}

\author{
Quansheng Hu\inst{1} \and
Qin Sun\inst{1} \and
Qiansen Dai\inst{1} \and
Jin Ding\inst{1} \and 
\\
Wan Zhang\inst{1} \and
Xue Zhou\inst{2,1}\thanks{Corresponding author} \and
Jianxiao Zou\inst{2}
}
\authorrunning{Q.~Hu et al.}

\institute{University of Electronic Science and Technology of China, Chengdu, China
\email{\{huhansan,sunqin,daiqiansen\}@std.uestc.edu.cn}, \email{zhouxue@uestc.edu.cn} 
\and
Shenzhen Institute for Advanced Study, UESTC, Shenzhen, China
}
\maketitle
\begin{abstract}
Referring multi-object tracking (RMOT) aims to track every instance in a video that matches a given language expression.
Despite the recent integration of multimodal large language models (MLLMs) to enhance generalization, existing methods predominantly relegate them to the role of caption generators, necessitating external modules for final decision-making. 
This paradigm not only introduces extra latency but also severely underutilizes the inherent vision–language alignment capabilities of MLLMs.
To address these limitations, we propose YesTrack, a novel two-stage RMOT method that reformulates referring as a discriminative task, directly leveraging MLLMs for Yes/No verification without explicit text generation. To further enhance the reliability and efficiency of this MLLM-based verification, we introduce two lightweight temporal consistency constraints: Temporal Confidence Prior (TCP) and Temporal Reference Propagation (TRP).
We further validate the generality of this discriminative paradigm by proposing YesTrack-MOT, a straightforward yet highly effective instantiation for generic multi-object tracking (MOT).
Experiments on Refer-KITTI and Refer-KITTI-V2 show that YesTrack significantly outperforms existing state-of-the-art methods while maintaining high efficiency, even when implemented with the smallest variant of Qwen3-VL. Code is released at \url{https://github.com/ggbondrighthere24/YesTrack}.
\keywords{Referring multi-object tracking \and Multimodal large language models \and Discriminative paradigm}
\end{abstract}

\section{Introduction}
Referring multi-object tracking (RMOT) is a multimodal video understanding task that aims to continuously localize and associate target objects across frames, conditioned on a natural language referring expression. Unlike generic multi-object tracking (MOT) \cite{bewley2016simple,wojke2017simple,zhang2022bytetrack,zeng2022motr,fang2025associate}, which tracks all instances indiscriminately, RMOT requires aligning specific trajectories with referring expressions, making it both practically relevant for applications like autonomous driving and a challenging benchmark for multimodal grounding in dynamic scenes.


\begin{figure}[tb]
  \centering
  \begin{subfigure}{0.6\linewidth}
    \includegraphics[width=\textwidth]{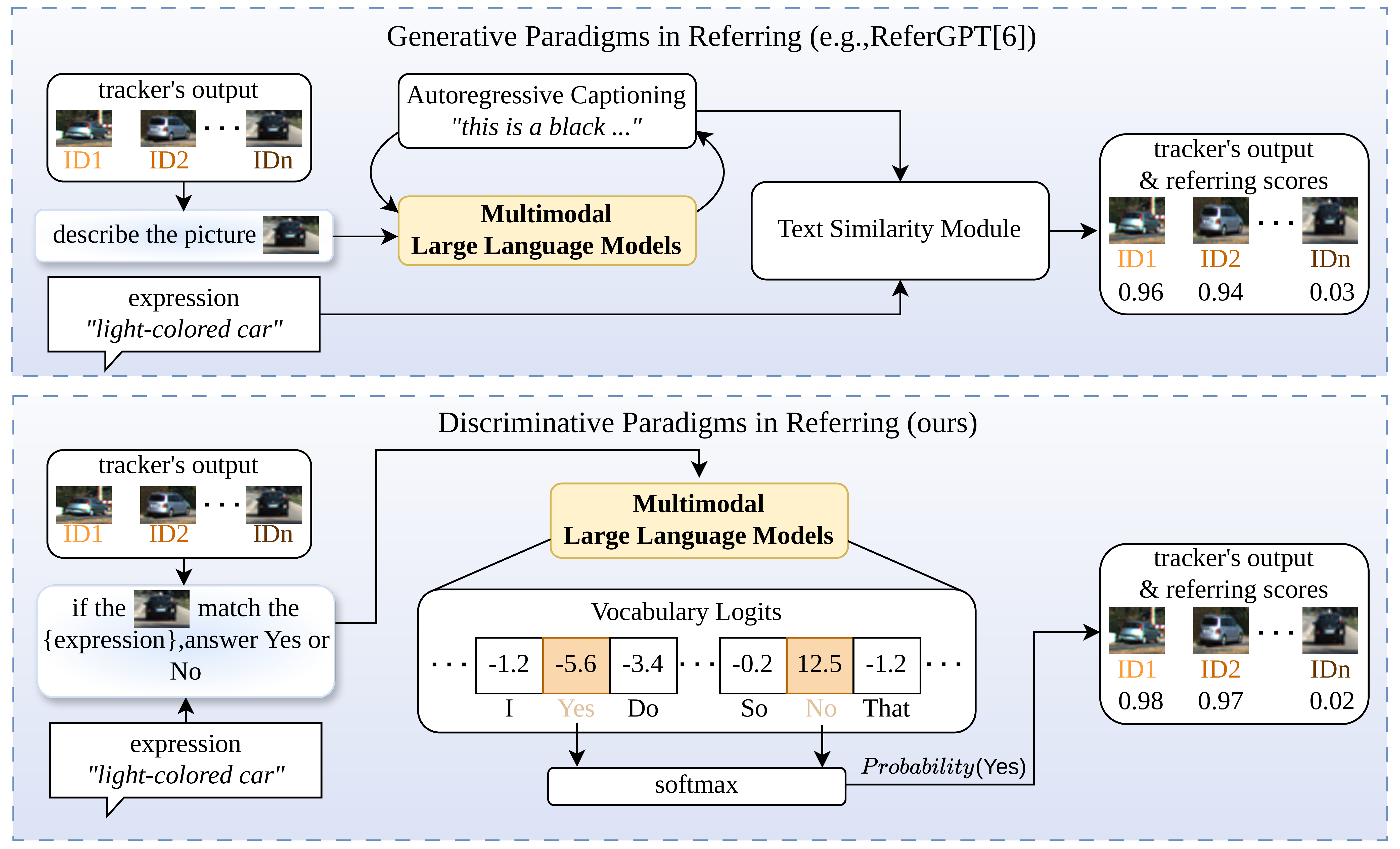}
    \caption{Generative vs. discriminative paradigms in referring.}
    \label{firstpart1}
  \end{subfigure}
  \hfill
  \begin{subfigure}{0.38\linewidth}
    \includegraphics[width=\textwidth]{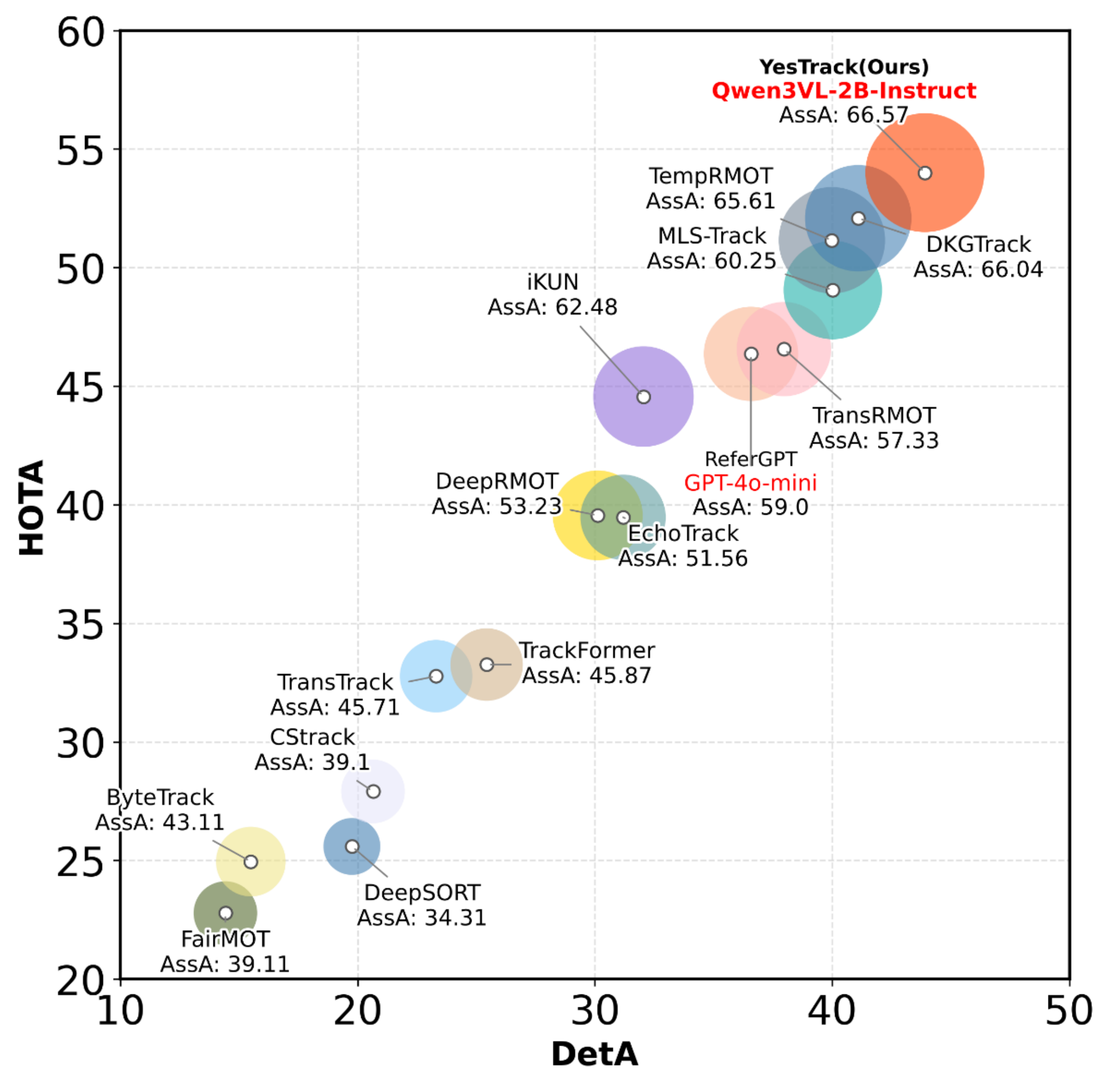}
    \caption{Performance comparison.}
    \label{firstpart2}
  \end{subfigure}
  \caption{(a) Generative methods first produce captions in an autoregressive manner and then apply an additional text–text similarity module for alignment, introducing latency from both sequential decoding and extra textual matching.
    Our discriminative paradigm directly extracts the intermediate logits from MLLMs and applies a softmax operation only over the logits corresponding to \emph{Yes} and \emph{No}, enabling a more efficient and direct use of multimodal alignment.
    (b) Our approach achieves state-of-the-art results with HOTA 54.00, DetA 43.91, and AssA 66.57 on Refer-KITTI.}
  \label{fig:short}
\end{figure}

Existing RMOT approaches mainly rely on transformer-based architectures to model cross-modal interactions and temporal dependencies. Most end-to-end methods \cite{li2025language,Zhao_Hao_Zhang_Liu_Li_Sui_He_Chen_2025} achieve strong performance with task-specific designs, such as decomposing expressions, fine-grained linguistic parsing, or multimodal attention. However, these methods are typically trained on datasets with limited vocabulary and expression diversity, which may introduce dataset-dependent biases and limit generalization to more complex or real-world scenarios.
To improve generalization, recent work \cite{Chamiti_2025_CVPR} formulate RMOT as a two-stage pipeline consisting of a tracking backbone followed by a language-conditioned referring module. In this paradigm, multimodal large language models (MLLMs) \cite{bai2025qwen25vltechnicalreport,bai2025qwen3vltechnicalreport,gemmateam2025gemma3technicalreport,liu2023visualinstructiontuning,wang2025internvl35advancingopensourcemultimodal} are primarily introduced in the referring stage to enhance cross-modal reasoning capability, often employed as caption generators, which is illustrated in Fig.~\ref{firstpart1}. While this enhances robustness to diverse expressions, autoregressive decoding introduces latency, and extra modules are required to convert textual outputs into concrete referring decisions, increasing architectural complexity and hindering tracking efficiency.

To address these limitations, we propose \textbf{YesTrack}, a simple yet effective RMOT framework that introduces a fundamentally different way of leveraging MLLMs. As illustrated in Fig.~\ref{firstpart1}, rather than employing MLLMs for caption generation, YesTrack adopts a discriminative paradigm to exploit MLLMs for referring.
Specifically, we reformulate referring as a binary image–text matching problem and leverage the MLLMs' output to directly determine whether a candidate trajectory matches the given referring expression.
This design completely avoids autoregressive decoding, requires only a single forward pass per candidate, and eliminates the need for additional complex for referring, thereby enabling efficient and
scalable inference.
Moreover, we introduce two temporal consistency constraints to further enhance robustness and efficiency for two-stage RMOT methods.
Temporal Confidence Prior (TCP) exploits identity-level temporal consistency to stabilize referring decisions under occlusion.
Temporal Reference Propagation (TRP) accelerates inference by performing MLLMs verification on sparsely sampled frames and propagating decisions to subsequent frames, while still handling newly appearing identities in a timely manner.
Moreover, this discriminative paradigm can be naturally extended to multi-object tracking by leveraging MLLMs as a pairwise identity verifier to replace traditional embedding-based ReID in data association, offering a straightforward yet effective way to incorporate MLLMs into key MOT components.

Our main contributions are summarized as follows:

(1) We propose YesTrack, a simple yet effective framework that unlocks the latent discriminative power of MLLMs for RMOT, transforming them from caption generators into discriminative referring heads. We further show that this discriminative paradigm generalizes beyond referring tasks—instantiated as YesTrack-MOT for generic multi-object tracking—achieving strong performance and confirming that direct MLLMs-driven decision-making can serve as a versatile foundation for tracking. 

(2) We introduce two lightweight yet general temporal consistency constraints for two-stage RMOT, namely Temporal Confidence Prior and Temporal Reference Propagation, to mitigate frame-wise prediction instability and reduce redundant cross-modal verification, thereby improving both robustness and computational efficiency.

(3) As summarized in Fig.~\ref{firstpart2}, extensive experiments on Refer-KITTI and Refer-KITTI-V2 demonstrate that YesTrack achieves state-of-the-art results with strong generalization and high efficiency, even when instantiated with Qwen3-VL-2B-Instruct, the smallest model in the Qwen3-VL family.

\section{Related Work}

\subsection{Referring Multi-object Tracking}
Referring multi-object tracking was first formalized by TransRMOT \cite{wu2023referring}, which introduced the task definition and the Refer-KITTI benchmark, framing RMOT as a joint problem of referring expression comprehension \cite{Deruyttere_2019,khoreva2019videoobjectsegmentationlanguage,nagaraja2016modelingcontextobjectsreferring,yu2016modelingcontextreferringexpressions} and multi-object tracking \cite{bewley2016simple,wojke2017simple,zhang2022bytetrack,zeng2022motr,fang2025associate}.
Existing methods fall into two paradigms: end-to-end and two-stage.
End-to-end methods integrate tracking and referring into a unified framework. TempRMOT \cite{zhang2024bootstrapping} introduces temporal enhancement for improved sequence modeling and releases the Refer-KITTI-V2 benchmark; DKGTrack \cite{li2025language} decomposes expressions into static and motion cues for fine-grained alignment; HFF-Tracker \cite{Zhao_Hao_Zhang_Liu_Li_Sui_He_Chen_2025} uses hierarchical feature fusion with adaptive training. Despite strong results, many end-to-end approaches rely on task-specific modules and training over a restricted vocabulary space, potentially limiting generalization to unseen scenarios.
Two-stage methods decouple tracking and referring, performing generic MOT first and then selecting target trajectories via language. Representative examples include iKUN \cite{du2024ikun}, which adds a Knowledge Unification Module to align visual and language features and Neural Kalman Filter to improve temporal association, and ReferGPT \cite{Chamiti_2025_CVPR}, which incorporates MLLMs to generate captions aligned with referring expressions. While effective, these sequential pipelines increase computational overhead and inference latency.

\subsection{MLLMs in Vision Language Understanding}

Recent works have incorporated multimodal large language models (MLLMs) into vision–language understanding such as person re-identification (ReID) \cite{ye2021deeplearningpersonreidentification,hermans2017defensetripletlossperson,luo2019bagtricksstrongbaseline,he2021transreidtransformerbasedobjectreidentification,wang2019rgb}. 
A prevalent paradigm treats MLLMs as generative description modules: visual content is first converted into textual descriptions and then aligned with textual queries for matching. This strategy is adopted in text-to-image ReID frameworks such as CLIP-SCGI \cite{han2024clip}, and in the referring stage of two-stage RMOT methods like ReferGPT \cite{Chamiti_2025_CVPR}, where caption generation and query matching are performed sequentially.
Another line of work exploits MLLMs for semantic token generation or attribute extraction, e.g., LVLM-ReID \cite{wang2024large}, which produces compact pedestrian representations to guide identity learning. VQA-style formulations have also been explored, treating ReID as an interactive reasoning process, as in LLaVA-ReID \cite{lu2025llava}. 

Although these approaches demonstrate the benefits of semantic enrichment and cross-modal alignment, applying generative paradigms directly to RMOT encounters a critical bottleneck: the autoregressive generation of numerous text tokens incurs significant latency, severely hindering the real-time inference required for tracking. To bypass this token generation overhead, recent practices that recast large models into discriminative roles offer valuable inspiration. By reducing complex multimodal reasoning to efficient binary Yes/No decisions, discriminative formulations provide a much faster alternative. 
Such discriminative formulations have been increasingly adopted as lightweight decision mechanisms, where large models are prompted to produce explicit Yes/No judgments for evaluation, task verification, and iterative control~\cite{shinn2023reflexionlanguageagentsverbal,tian2025macgyverlargelanguagemodels,sun2024thinkongraphdeepresponsiblereasoning}. These practices suggest that, beyond generation, large models possess strong discriminative potential; however, existing approaches essentially treat the model as a binary classifier by directly eliciting discrete Yes/No outputs, without leveraging the underlying token-level logits, which limits their effectiveness for RMOT.

\section{Method}
In this section, we describe the proposed YesTrack framework.
YesTrack is a two-stage RMOT paradigm that combines an off-the-shelf tracker with an MLLM-based discriminative referring module.
On top of this basic pipeline, we introduce two intuitive yet general temporal consistency constraints to improve robustness and efficiency.
Beyond the referring stage in RMOT, we present a tracker, termed YesTrack-MOT, which applies the same discriminative MLLMs paradigm to data association in MOT.
\subsection{Framework Overview}
\begin{figure}[t]
\centering
\includegraphics[width=1\textwidth]{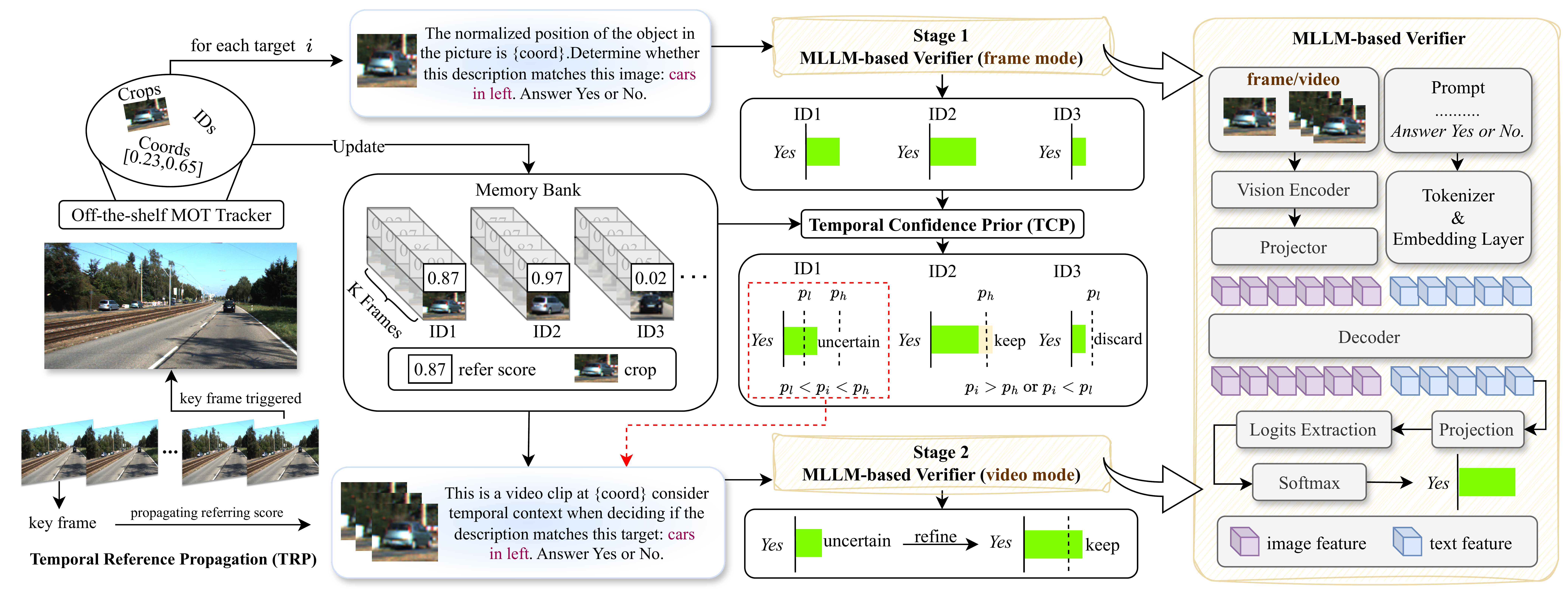}
\caption{Overview of YesTrack. An off-the-shelf tracker extracts candidate crops, and Temporal Reference Propagation (TRP) propagates recent decisions. Remaining candidates are handled by a two-stage MLLM-based verifier: Stage 1 performs single-frame scoring with Temporal Confidence Prior (TCP) regularization for early routing, while Stage 2 refines uncertain cases at the video level using a memory bank. The right panel shows the verifier architecture, where frame and video modes differ only in input image number and prompt formulation. Final binary matching probabilities are obtained by applying softmax to the projected decision token logits.
}
\label{fig:wide-example}
\end{figure}

As illustrated in Fig.~\ref{fig:wide-example}, given a video and a referring expression, YesTrack first employs Temporal Reference Propagation (TRP) to reduce redundant verification by leveraging the temporal stability of referring decisions. Specifically, TRP categorizes frames into key and non-key frames: a frame is defined as a key frame if MLLM-based referring is triggered at that frame (see \cref{sec:tcc}). At key frames, full verification is performed, while for non-key frames, the previously computed referring score is directly propagated for each active identity. This mechanism avoids unnecessary MLLM invocations across consecutive frames. To facilitate this process, YesTrack applies an off-the-shelf tracker to each frame $t$ to obtain the candidate targets indexed by $i$. For each target $i$, the tracker provides its track crop, identity ID, and normalized bounding box coordinates.
For candidates requiring verification at key frames, we employ a discriminative two-stage MLLM-based verifier. In Stage 1 (frame mode), the MLLM performs single-frame pairwise verification between the crop of target $i$ and the referring expression, producing a confidence score $p_i \in [0,1]$ for each candidate. Temporal Confidence Prior (TCP) regularization is then applied to incorporate historical confidence signals and stabilize predictions (\cref{sec:tcc}).
Based on the TCP-adjusted confidence $p_i$, candidates are routed efficiently: those with $p_i \geq p_h$ or $p_i \leq p_l$ (where $p_h$ and $p_l$ denote predefined upper and lower confidence thresholds) are directly kept or discarded, respectively. Ambiguous candidates satisfying $p_l < p_i < p_h$ proceed to Stage~2.
In Stage 2 (video mode), the verifier performs temporal refinement. A memory bank retrieves the past $K$ frames associated with target $i$, including their corresponding crops and referring scores, where $K$ denotes the temporal window size. The MLLMs jointly evaluates this temporal context and outputs the final discriminative decision, which is determined by a threshold $\gamma$.

\subsection{MLLM-based Verifier}
\label{mllmverifier}

To determine whether a target instance matches the given referring expression, we adopt MLLMs as a binary verifier.
The verifier receives a cropped target image or cropped video clip together with the
referring expression and outputs a binary decision indicating
whether the candidate matches the description.

\subsubsection{Input formatting.}
For each candidate identity proposed by the tracker, we construct a unified multimodal input consisting of the cropped image region, its normalized spatial coordinates, and the referring expression.
The spatial coordinates are embedded into the textual prompt
to provide explicit location cues.

To constrain the output space, we explicitly instruct the model
to answer \emph{Yes} or \emph{No}.
In contrast to generative approaches where the model is free to
produce variable-length text, this binary formulation restricts
the decision space to two semantic options.
Such a closed-set setup encourages sharper decision boundaries
in the output logits and reduces ambiguity introduced by
open-ended decoding.
It also aligns naturally with the instruction-following
capabilities of modern MLLMs, enabling stable and efficient
inference.

Separate prompt templates are used for frame mode and
video mode while maintaining the same binary
decision format.
The exact templates are provided in the supplementary material.

\subsubsection{Binary probability extraction.}
As illustrated in Fig.~\ref{fig:wide-example}, after jointly encoding the visual and textual inputs, MLLMs produce token-level logits over the vocabulary after decoding and projection. 
Instead of relying on autoregressive decoded text, we directly extract the logits corresponding to the two decision tokens.
Given an input pair $(I_i, E)$ for tracked identity $i$, where $I_i$ is the image crop and $E$ is the corresponding expression, we denote the logits of the decision tokens as $\ell_i^{\text{yes}}$ and $\ell_i^{\text{no}}$, and compute the matching probability of \emph{Yes} $p_i$ using a softmax function:
\begin{equation}
p_i
=
\frac{\exp(\ell^{\text{yes}}_i)}
     {\exp(\ell^{\text{yes}}_i) + \exp(\ell^{\text{no}}_i)}.
\end{equation}
Compared with generative decoding, this logit-based formulation provides several distinct advantages. By bypassing free-form outputs, it inherently avoids the instability associated with unexpected explanations or wording variations. Furthermore, the extracted logits translate directly into a continuous confidence score. This preserves the uncertainty information essential for confidence-based routing and temporal refinement. Consequently, this design eliminates the computational overhead of iterative token sampling, streamlining the process to require only a single forward pass per candidate.

\subsubsection{Training.}
Following the binary probability formulation above, each input pair $(I_i,E)$ is annotated with a binary ground-truth label $y_i \in \{0, 1\}$, where $y_i = 1$ indicates a positive match and $y_i = 0$ otherwise. 
Utilizing the discriminative matching probability $p_i$ computed from the two decision logits, the training objective is formulated as the standard Binary Cross-Entropy (BCE) loss:

\begin{equation}
\mathcal{L}_i = - \left[ y_i \log p_i + (1 - y_i) \log (1 - p_i) \right],
\end{equation}
where the softmax normalization is restricted to the two decision tokens.

\subsection{Temporal Consistency Constraints}
\label{sec:tcc}
To improve both robustness and efficiency in this two-stage RMOT pipeline, we introduce two simple inference-time temporal consistency constraints: Temporal Reference Propagation (TRP) and Temporal Confidence Prior (TCP).
These strategies enhance temporal robustness without modifying the core model architecture and can be readily extended to other two-stage RMOT frameworks.

\subsubsection{Temporal Reference Propagation (TRP).}

TRP is motivated by the observation that referring decisions are typically more temporally stable than tracking states. While trajectories require frame-wise updates, the semantic relevance of an identity to a given query rarely changes within a short temporal window. Exploiting this property enables us to reduce redundant MLLM invocations without sacrificing referring accuracy.

Specifically, a frame is defined as a key frame if MLLM-based referring is triggered at that frame. At each key frame, both tracking and referring are performed, and the resulting referring score is propagated to subsequent frames until the next key frame. For non-key frames, only tracking is executed, while the referring score is directly inherited from the most recent key frame as long as the identity remains active. In practice, referring is triggered at a fixed interval $\Delta$, when a new identity appears, or when the referring expression changes.

\subsubsection{Temporal Confidence Prior (TCP).}

TCP exploits identity-level temporal consistency across frames.
For each tracked identity $i$ at frame $t$,
let $p_i^{(t)} \in [0,1]$ be the matching probability
predicted by the MLLMs.
If an identity has been confidently matched in recent frames,
it is likely to remain relevant in the current frame.
Based on this observation, TCP introduces a temporal prior
to bias the current prediction.
The adjusted probability is formulated as follows:


\begin{equation}
\tilde{p}_i^{(t)}
=
\min\left\{
1,\,
p_i^{(t)}
+
\lambda \cdot
\mathbb{I}\left(
\min_{k \in [t-K,\, t-1]} p_i^{(k)} \ge \alpha
\right)
\right\},
\end{equation}
where $\mathbb{I}(\cdot)$ denotes the indicator function,
$K$ is the temporal window size,
$\alpha$ is a confidence threshold,
and $\lambda > 0$ controls the strength of the temporal prior.
The outer $\min(\cdot)$ clips the adjusted score to 1, ensuring that
$\tilde{p}_i^{(t)}$ remains a valid probability.
The inner $\min(\cdot)$ operator enforces strict temporal consistency:
the prior is applied only when the identity remains confidently
matched across all previous $K$ frames.

\subsection{YesTrack-MOT}
While YesTrack is a two-stage RMOT framework that consists of a tracking backbone
and a subsequent MLLM-based referring module, its tracking component can be
implemented by any off-the-shelf MOT tracker.

\begin{figure*}[t] 
\centering
\includegraphics[width=1\textwidth]{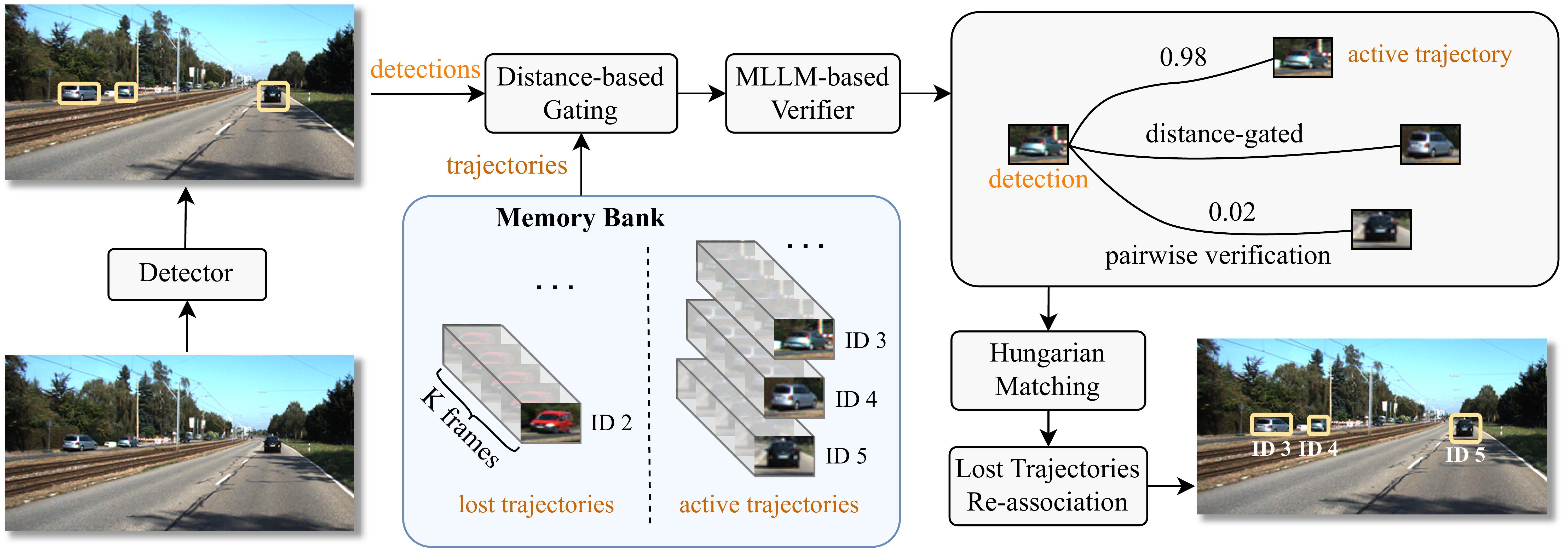}
\caption{Overview of YesTrack-MOT. Given detections from a standard detector, YesTrack-MOT first applies distance-based gating to reduce candidate pairs. An MLLM-based verifier then performs pairwise image–image verification between current detections and previous track crops, producing a confidence score that forms a cost matrix. Tracks are updated via Hungarian matching, while unmatched tracks are temporarily kept for potential re-association, forming a minimal yet effective MOT pipeline.}
\label{fig:network}
\end{figure*}

In this work, we further instantiate the tracking backbone with a simple
MLLM-based tracker, termed YesTrack-MOT, by extending the same
discriminative paradigm to data association. As illustrated in Fig.~\ref{fig:network}, YesTrack-MOT replaces conventional embedding-based ReID with an
MLLM-based pairwise identity verifier, resulting in a deliberately
simple MOT instantiation.

Importantly, the underlying idea is consistent with our referring module:
we treat the MLLMs as a verifier that outputs a continuous confidence score.
The only difference lies in the input modality: referring performs
image--text verification, whereas YesTrack-MOT performs image--image
verification between a tracked crop and a detection crop.
For clarity, the exact prompt templates used for identity verification are
provided in the supplementary material.

\subsubsection{Simple distance gating.}
At frame $t$, let $\mathcal{T}_{t-1}=\{ \tau_j \}_{j=1}^{N}$ be the set of
tracks from the previous frame and $\mathcal{D}_{t}=\{ d_k \}_{k=1}^{M}$ be the detections in the current frame.
To reduce the number of candidate pairs, we apply a lightweight distance-based gating rule and only keep plausible pairs whose center displacement is within a threshold:
\begin{equation}
g(\tau_j, d_k)=
\mathbb{I}\left(
\left\|c(\tau_j)-c(d_k)\right\|_2 \le \delta
\right),
\end{equation}
where $c(\cdot)$ denotes the bounding-box center, $\delta$ is a gating
threshold and $\mathbb{I}(\cdot)$ denotes the indicator function.

\subsubsection{MLLM-based identity verification.}
For each gated pair $(\tau_j, d_k)$ such that  
$g(\tau_j,d_k)=1$, we crop the corresponding image regions from frames $t-1$ and $t$, denoted as $I(\tau_j)$ and $I(d_k)$, respectively. 
Consistent with the probability extraction introduced in \cref{mllmverifier}, we formulate the identity verification as a purely discriminative task. The underlying mechanism remains identical, adopting MLLMs as a binary verifier. The only difference lies in the input modality, where the MLLMs now processes two visual inputs jointly rather than an image-text pair. 
Instead of relying on autoregressive text decoding, we directly extract the logits corresponding to the binary decision tokens. Let $\ell_{jk}^{\text{yes}}$ and $\ell_{jk}^{\text{no}}$ denote the extracted logits for the positive and negative verification outcomes of the input pair $(I(\tau_j), I(d_k))$. The verification probability $p_{jk}$ is then computed using a softmax function:
\begin{equation}
p_{jk} = \frac{\exp(\ell^{\text{yes}}_{jk})}{\exp(\ell^{\text{yes}}_{jk}) + \exp(\ell^{\text{no}}_{jk})}.
\end{equation}

\subsubsection{Hungarian matching with simple track management.}
We construct a cost matrix $C = [c_{jk}]$ from the verification probabilities and solve a
one-to-one assignment using the Hungarian algorithm \cite{kuhn1955hungarian}:
\begin{equation}
c_{jk} = 1 - p_{jk}, \qquad \pi^\star = \arg\min_{\pi} \sum_{(j,k)\in \pi} c_{jk},
\end{equation}
where $\pi$ denotes all feasible matching between tracks and detections under the gated candidate set and $\pi^\star$ denotes the optimal assignment.
Matched pairs update the corresponding tracks, while unmatched tracks are
marked as lost.
To handle short-term occlusions, lost tracks are kept for the subsequent $L$
frames; during this period, unmatched detections are additionally matched
against lost tracks once for potential re-association, otherwise a track is removed after exceeding the maximum lost age.

Overall, YesTrack-MOT intentionally avoids sophisticated motion models and builds a minimal MOT pipeline with only distance
gating, MLLM-based verification, and Hungarian assignment, yet still achieves
strong performance in practice.

\section{Experiment}

\subsection{Datasets and Implementation Details}

\subsubsection{Datasets.}
We evaluate our method on two RMOT benchmarks: Refer-KITTI \cite{wu2023referring} and Refer-KITTI-V2 \cite{zhang2024bootstrapping}.
Refer-KITTI, the first dataset for RMOT, is built from 18 KITTI tracking sequences and contains 818 expressions, among which only 215 are distinct, with a vocabulary of 49 words. The expressions are generally short, repetitive, and structurally simple, and the limited subset of sequences leads to restricted scene and motion diversity.
In contrast, Refer-KITTI-V2 substantially expands the dataset to all 21 sequences and introduces 9,758 expressions, including 7,193 distinct ones with a vocabulary of 617 words. Compared to the original version, Refer-KITTI-V2 exhibits significantly richer semantics, covering appearance, spatial relations, motion, and inter-object interactions, while also incorporating ambiguous expressions and no-target cases. Overall, Refer-KITTI-V2 presents a much larger linguistic space and more challenging scenarios, making it a more realistic and demanding benchmark for evaluating RMOT methods.

\begin{table*}[t]
\centering
\caption{Comparison with state-of-the-art RMOT methods on the Refer-KITTI dataset. We group methods into end-to-end and two-stage pipelines. ReferGPT(Q) denotes using Qwen3-VL-2B as the backbone, consistent with our method, while ReferGPT(O) denotes its original MLLM configuration. The best results are highlighted in \textbf{bold}.}
\label{tab:referkitti}
\resizebox{\textwidth}{!}{
\begin{tabular}{l c cccccccc}
\hline
Methods & Tracker & HOTA $\uparrow$ & DetA $\uparrow$ & AssA $\uparrow$ & DetRe $\uparrow$ & DetPr $\uparrow$ & AssRe $\uparrow$ & AssPr $\uparrow$ & LocA $\uparrow$ \\
\hline
\multicolumn{10}{l}{\textbf{End-to-End}} \\
\hline
EchoTrack \cite{lin2024echotrack}  & -- & 39.47 & 31.19 & 51.56 & 42.65 & 48.86 & 56.68 & 81.21 & 79.93 \\
DeepRMOT \cite{he2024visual} & -- & 39.55 & 30.12 & 53.23 & 41.91 & 47.47 & 58.47 & 82.16 & 80.49 \\
TransRMOT \cite{wu2023referring} & -- & 46.56 & 37.97 & 57.33 & 49.69 & 60.10 & 60.02 & 89.67 & 90.33 \\
MGLT-MOTR \cite{chen2025multigranularity} & -- & 47.95 & 40.04 & 57.57 & --    & --    & --    & --    & --    \\
MLS-Track \cite{ma2024mls} & -- & 49.05 & 40.03 & 60.25 & 59.07 & 54.18 & 65.12 & 88.12 & --    \\
CDRMT \cite{liang2025cognitive}  & -- & 49.35 & 40.34 & 60.56 & 54.54 & 59.30 & 64.70 & \textbf{89.80} & 90.61 \\
TenRMOT \cite{xiao2025temporal}  & -- & 49.77 & 40.79 & 60.89 & 52.65 & \textbf{62.81} & 65.38 & 89.28 & \textbf{90.69} \\
TempRMOT \cite{zhang2024bootstrapping} & -- & 51.15 & 39.99 & 65.61 & 54.23 & 58.89 & 70.91 & 87.27 & 90.46 \\
DKGTrack \cite{li2025language} & -- & 52.08 & 41.10 & 66.04 & 57.57 & 58.36 & 71.13 & 87.98 & 90.54 \\
\hline
\multicolumn{10}{l}{\textbf{Two-stage}} \\
\hline
iKUN \cite{du2024ikun}  & NeuralSORT \cite{du2024ikun} & 44.56 & 32.05 & 62.48 & 48.53 & 44.76 & 70.52 & 76.66 & -- \\
MEX  \cite{tran2024mex}  & NeuralSORT \cite{du2024ikun} & 45.07 & 32.81 & 62.52 & 54.84 & 41.65 & 71.09 & --    & --    \\
ReferGPT(Q) \cite{Chamiti_2025_CVPR}  & PC3T \cite{PC3T}  & 32.56 & 17.41 & 61.14 & 26.39 & 32.23 & 71.80 & 71.99 & 81.67 \\
ReferGPT(O) \cite{Chamiti_2025_CVPR}  & PC3T \cite{PC3T}  & 46.36 & 36.58 & 59.00 & 51.40 & 52.16 & \textbf{73.16} & 69.31 & 83.26 \\
\hline
\textbf{YesTrack} (ours) & \textbf{TempRMOT\textsuperscript{$\ast$}} & \textbf{54.00} & 43.91 & \textbf{66.57} & 59.11 & 60.65 & 72.64 & 85.08 & 88.52 \\
\textbf{YesTrack} (ours) & \textbf{YesTrack-MOT} & 52.96 & \textbf{46.84} & 60.03 & \textbf{62.72} & 62.47 & 65.01 & 87.69 & 89.01 \\
\hline
\end{tabular}}
\end{table*}

\subsubsection{Implementation Details.}
We adopt Qwen3-VL-2B-Instruct \cite{bai2025qwen3vltechnicalreport} as the MLLMs backbone for referring. 
For each candidate bounding box, the cropped image region is resized to a resolution of $320 \times 320$ before being fed into the vision encoder.
In the frame mode, the binary matching probability is obtained from the logits of the \textit{Yes} and \textit{No} tokens. 
We use a confidence interval $[p_l, p_h]$ with $p_l = 0.2$ and $p_h = 0.8$ to determine prediction certainty. 
Predictions with probabilities within this interval are considered uncertain and further processed by the video mode. 
The final decision threshold $\gamma$ is set to $0.4$.
For Temporal Confidence Prior, the temporal window size $K$ is set to 3, 
the high-confidence threshold $\alpha$ is set to 0.4, 
and the temporal prior strength $\lambda$ is set to 0.3.
For Temporal Reference Propagation, the propagation interval $\Delta$ is set to 5 on Refer-KITTI and 10 on Refer-KITTI V2. 
The memory bank size is set to 4. In YesTrack-MOT, the distance threshold is set to 200 pixels and lost tracks are kept for 10 frames.
All experiments are conducted on two NVIDIA RTX 4090 GPUs with 24GB memory each.

\subsection{Evaluation Metrics}

Referring multi-object tracking is typically evaluated using standard MOT metrics, including Higher Order Tracking Accuracy (HOTA) \cite{luiten2021hota}, DetA, and AssA. Computed against the expression-specific ground-truth trajectory, these metrics jointly assess both referring correctness and temporal tracking consistency. However, for two-stage RMOT pipelines utilizing varying external trackers, standard MOT metrics cannot isolate the model's intrinsic referring capability from the underlying tracking variances.

To enable a fair, tracker-agnostic evaluation of cross-modal referring ability, we introduce binary referential metrics: Accuracy, Precision, and Recall. By adopting ground-truth tracklets as candidates, we eliminate tracking algorithm variance entirely. Formally, given $N$ candidates with ground-truth labels $y_i \in \{0,1\}$ and model predictions $\hat{y}_i$, the referential accuracy is defined as:
\begin{equation}
\mathrm{Acc}_{\mathrm{ref}}
=
\frac{1}{N}
\sum_{i=1}^{N}
\mathbb{I}\left(\hat{y}_i = y_i\right),
\end{equation}
where $\mathbb{I}(\cdot)$ is the indicator function. Precision and Recall follow standard definitions for the positive class, directly measuring target identification accuracy independent of tracking quality.

\begin{table*}[t]
\centering
\caption{Comparison with state-of-the-art RMOT methods on the Refer-KITTI-V2 dataset. We group methods into single-stage (end-to-end) and two-stage pipelines. The best results are highlighted in \textbf{bold}.}
\label{tab:referkittiv2}
\resizebox{\textwidth}{!}{
\begin{tabular}{l c cccccccc}
\hline
Methods & Tracker & HOTA $\uparrow$ & DetA $\uparrow$ & AssA $\uparrow$ & DetRe $\uparrow$ & DetPr $\uparrow$ & AssRe $\uparrow$ & AssPr $\uparrow$ & LocA $\uparrow$ \\
\hline
\multicolumn{10}{l}{\textbf{End-to-End}} \\
\hline
TransRMOT \cite{wu2023referring} & -- & 31.00 & 19.40 & 49.68 & 36.41 & 28.97 & 54.59 & 82.29 & 89.82 \\
TempRMOT  \cite{zhang2024bootstrapping} & -- & 34.72 & 22.52 & 53.64 & 32.41 & 41.76 & 58.98 & 83.16 & 90.38 \\
DKGTrack \cite{li2025language}  & -- & 35.26 & 23.04 & 54.13 & 37.81 & 36.88 & 60.73 & 83.85 & \textbf{91.65} \\
\hline
\multicolumn{10}{l}{\textbf{Two-stage}} \\
\hline
iKUN \cite{du2024ikun}   & NeuralSORT \cite{du2024ikun} & 10.32 & 2.17  & 49.77 & 2.36  & 19.75 & 58.48 & 68.64 & 74.56 \\
ReferGPT \cite{Chamiti_2025_CVPR}  & PC3T \cite{PC3T} & 30.12 & 15.69 & \textbf{59.02} & 21.55 & 34.41 & \textbf{74.59} & 68.20 & 79.76 \\
\hline
\textbf{YesTrack} (ours) & \textbf{TempRMOT\textsuperscript{$\ast$}} & 41.78 & 32.69 & 53.75 & 41.37 & \textbf{58.10} & 59.98 & 81.07 & 87.17 \\
\textbf{YesTrack} (ours) & \textbf{YesTrack-MOT} & \textbf{43.75} & \textbf{37.04} & 52.36 & \textbf{48.78} & 56.89 & 57.49 & \textbf{83.95} & 85.06 \\
\hline
\end{tabular}}
\end{table*}

\subsection{Benchmark Experiment}
As shown in Table \ref{tab:referkitti} and Table \ref{tab:referkittiv2}, our proposed YesTrack consistently delivers strong gains across both benchmarks. On Refer-KITTI, YesTrack with TempRMOT\textsuperscript{$\ast$} achieves the best overall HOTA 54.00 and the best association accuracy AssA 66.57, surpassing the strongest prior end-to-end method. With the YesTrack-MOT tracker, we further obtain the best DetA 46.84 and the top DetRe, showing that our framework improves both association quality and detection-related performance. On the more challenging Refer-KITTI-V2, YesTrack remains clearly superior: YesTrack-MOT attains the best HOTA 43.75 and DetA 37.04, and also leads on DetRe 48.78 and AssPr 83.95, while the TempRMOT\textsuperscript{$\ast$} variant still yields substantial improvements. Here TempRMOT\textsuperscript{$\ast$} is a pure MOT model constructed by removing all language inputs and multimodal fusion components from TempRMOT. We use it because many recent RMOT frameworks are built upon TempRMOT-style baselines, and adopting this stripped variant enables a more fair comparison by aligning the underlying tracker and isolating the effect of the referring strategy.

\subsection{Ablation Experiments}

\begin{table*}[t]
\centering
\caption{Ablation study of different inference modes and temporal refinement components on Refer-KITTI. Frame Mode and Video Mode denote frame-level and video-level referring inference, respectively. TCP and TRP represent two temporal refinement strategies applied during post-processing. The best results are highlighted in \textbf{bold}.}
\label{tab:ablation}
\scalebox{0.85}{
\begin{tabular}{cccc|cccc}
\toprule
Frame Mode & Video Mode & TCP & TRP & HOTA $\uparrow$ & DetA $\uparrow$ & AssA $\uparrow$ & Inference Time $\downarrow$\\
\midrule
\checkmark &  &  &  & 52.64  & 42.68 & 65.06 & 40mins \\ 
& \checkmark &  &  & 53.49 & 43.29 & 66.24 & 2h15mins\\ 
\checkmark & \checkmark &  &  & 53.47  & 43.70  & 65.60 & 1h32mins\\
\checkmark & \checkmark & \checkmark &  & \textbf{54.00} & \textbf{43.91} & 66.57 & 1h34mins \\ 
\checkmark & \checkmark &  & \checkmark & 53.54 & 43.69 & 65.88 & \textbf{22mins} \\
\checkmark & \checkmark & \checkmark  & \checkmark & 53.66 & 43.27 & \textbf{66.71} & 25mins \\ 
\bottomrule
\end{tabular}}
\end{table*}

Table~\ref{tab:ablation} presents ablation studies on Refer-KITTI to analyze the effects of different inference modes and temporal refinement strategies. 
Using only frame mode already yields strong performance, while only video mode further improves HOTA and association metrics at the cost of increased inference time. Combining frame mode and video mode inference provides a better balance between detection and association quality, demonstrating their complementary roles.


Introducing TCP achieves the best overall HOTA and DetA, indicating that leveraging historical confidence across frames effectively stabilizes trajectory selection. In contrast, TRP slightly reduces association accuracy but dramatically accelerates inference, achieving the fastest runtime while maintaining competitive performance. Combining TCP and TRP yields the best AssA with only a small runtime increase over TRP alone. Notably, with TRP enabled, our method attains the highest inference speed among all ablation variants. Detailed comparisons of training and inference time against other methods are provided in the supplementary material. These results highlight a clear trade-off between accuracy and efficiency and confirm that the proposed temporal refinement components are effective on Refer-KITTI.


\subsection{Performance with Various Base Trackers}

Table~\ref{tab:trackers} presents a comprehensive analysis of different referring strategies applied to various trackers. We select four representative trackers for evaluation: (i) \textbf{ByteTrack} \cite{zhang2022bytetrack}, a widely used mature multi-object tracker; (ii) \textbf{TempRMOT$^\ast$} \cite{zhang2024bootstrapping}, a pure tracking baseline derived from TempRMOT; (iii) \textbf{YesTrack-MOT}, the visual tracking component of our proposed method; and (iv) \textbf{Ground Truth tracks}, which provide an upper-bound analysis. Crucially, for the referring strategy comparison, we select iKUN \cite{du2024ikun} as the primary baseline, as it represents the state-of-the-art performance among existing two-stage methods.

\begin{table}[t]
\centering
\caption{Performance comparison of different referring methods under various trackers on the Refer-KITTI dataset. MOT metrics evaluate tracking quality, while referential metrics evaluate language-based target selection. ``---'' denotes tracking without language guidance. The best results under each tracker are marked in \textbf{bold}.}
\label{tab:trackers}
\scalebox{0.9}{\begin{tabular}{c ccc c ccc}
\toprule
\multirow{2}{*}{\textbf{Method}} 
& \multicolumn{3}{c}{\textbf{Referential Metrics}} 
& \multirow{2}{*}{\textbf{Tracker}}
& \multicolumn{3}{c}{\textbf{RMOT Metrics}} \\
\cmidrule(lr){2-4} \cmidrule(lr){6-8}
& Acc $\uparrow$ & Prec. $\uparrow$ & Recall $\uparrow$ 
& 
& HOTA $\uparrow$ & DetA $\uparrow$ & AssA $\uparrow$ \\
\midrule

\multirow{4}{*}{---}
 & \multirow{4}{*}{--} & \multirow{4}{*}{--} & \multirow{4}{*}{--} 
 & ByteTrack \cite{zhang2022bytetrack} & 26.02 & 12.33 & \textbf{55.70} \\
 & & &
 & TempRMOT\textsuperscript{$\ast$} \cite{zhang2024bootstrapping}& 34.02 & 16.70 & \textbf{69.37} \\
 & & & 
 & YesTrack-MOT & 32.31 & 16.86 & \textbf{62.00} \\
 & & &
 & GT & 42.17 & 19.64 & \textbf{90.56} \\
\midrule

\multirow{4}{*}{iKUN \cite{du2024ikun}}
 & \multirow{4}{*}{84.62} & \multirow{4}{*}{59.88} & \multirow{4}{*}{70.24}
 & ByteTrack \cite{zhang2022bytetrack} & 33.59 & 22.30 & 51.24 \\
 & & &
 & TempRMOT\textsuperscript{$\ast$} \cite{zhang2024bootstrapping} & 42.95 & 28.11 & 65.65 \\
 & & & 
 & YesTrack-MOT & 45.81 & 35.80 & 58.71 \\
 & & &
 & GT & 55.97 & 37.93 & 82.58 \\
\midrule
\multirow{4}{*}{YesTrack}
 & \multirow{4}{*}{\textbf{91.15}} & \multirow{4}{*}{\textbf{74.14}} & \multirow{4}{*}{\textbf{85.71}}
 & ByteTrack \cite{zhang2022bytetrack}& \textbf{44.26} & \textbf{36.63} & 54.11 \\
 & & &
 & TempRMOT\textsuperscript{$\ast$} \cite{zhang2024bootstrapping}& \textbf{54.00} & \textbf{43.91} & 66.57 \\
 & & & 
& YesTrack-MOT & \textbf{52.96} & \textbf{46.84} & 60.03 \\
 & & &
 & GT & \textbf{72.80} & \textbf{62.17} & 85.25 \\
\bottomrule
\end{tabular}}
\end{table}

In terms of referential metrics, YesTrack demonstrates stronger capability in interpreting textual descriptions compared to iKUN, consistently outperforming it across all evaluation metrics, including Accuracy, Precision, and Recall. These significant gains indicate that our method is more effective at distinguishing target objects from distractors based on language cues, ensuring high-quality candidates for the subsequent tracking stage.
As expected, the no-referring baseline consistently achieves the highest AssA. Since referring methods merely filter existing trajectories based on language relevance rather than improving intrinsic identity association, their AssA is naturally upper-bounded by the underlying tracker.
Regarding RMOT Metrics, YesTrack consistently outperforms the iKUN baseline across all four underlying trackers.  This consistent improvement across varying tracker qualities demonstrates that YesTrack  possesses a stronger referring capability.

\begin{table*}[t]
\centering
\caption{Comparison of tracking-by-detection MOT methods on KITTI. 
Note that the same validation split as Refer-KITTI-V2 is used. 
Evaluation is conducted only on the car and person classes. The best results are highlighted in bold.}
\label{tab:referkittiv2mot}
\resizebox{\textwidth}{!}{
\begin{tabular}{l c cccccccc}
\hline
Methods &  HOTA $\uparrow$ & DetA $\uparrow$ & AssA $\uparrow$ & DetRe $\uparrow$ & DetPr $\uparrow$ & AssRe $\uparrow$ & AssPr $\uparrow$ & LocA $\uparrow$ \\
\hline
BoT-SORT \cite{aharon2022botsortrobustassociationsmultipedestrian}& 38.92 & 29.88 & 51.65 & 31.14 & \textbf{76.94} & 54.93 & \textbf{79.77} & 81.46\\
TrackTrack \cite{Shim_2025_CVPR} & 40.24 & 32.24 & 51.29 & 33.79 & 76.72 & 54.81 & 79.58 & \textbf{81.57}\\
ByteTrack \cite{zhang2022bytetrack}  & 40.83 & 32.85 & \textbf{52.30} & 34.60 & 73.02 & 55.84 & 76.91 & 78.96 \\
OC-SORT \cite{cao2023observationcentricsortrethinkingsort} & 41.34 & 34.64 & 50.57 & 36.79 & 74.56 & 54.26 & 78.67 & 80.97 \\
YesTrack-MOT (ours) & \textbf{45.10} & \textbf{40.96} & 51.23 & \textbf{47.59} & 64.52 & \textbf{55.99} & 77.79 & 79.12\\
\hline
\end{tabular}}
\end{table*}

\subsection{Comparison with MOT Methods}
To evaluate the tracking capability of YesTrack-MOT under the standard MOT setting, we compare it with several representative MOT methods. Since detection quality can significantly affect tracking performance, we standardize the detection inputs when comparing different methods. 
For a fair comparison, all tracking-by-detection methods in Table~\ref{tab:referkittiv2mot} use the same detection results generated by RF-DETR \cite{robinson2026rfdetrneuralarchitecturesearch}. In addition, BoT-SORT \cite{aharon2022botsortrobustassociationsmultipedestrian} and TrackTrack \cite{Shim_2025_CVPR} employ a FastReID \cite{he2020fastreidpytorchtoolboxgeneral} model trained on KITTI to provide appearance embeddings for data association.

It is worth noting that this comparison follows the standard MOT setting and does not involve referring expressions. As shown in Table~\ref{tab:referkittiv2mot}, despite adopting a relatively simple design, YesTrack-MOT achieves the best overall performance with a HOTA of 45.10, outperforming existing tracking-by-detection baselines. These results demonstrate that leveraging MLLMs as a verifier can provide effective cues for association, even without complex motion or appearance modeling. The strong performance suggests that treating MLLMs as a lightweight verification module is a promising alternative to conventional hand-crafted association strategies.

\subsection{Qualitative Analysis}
\begin{figure*}[t] 
\centering
\includegraphics[width=1\textwidth]{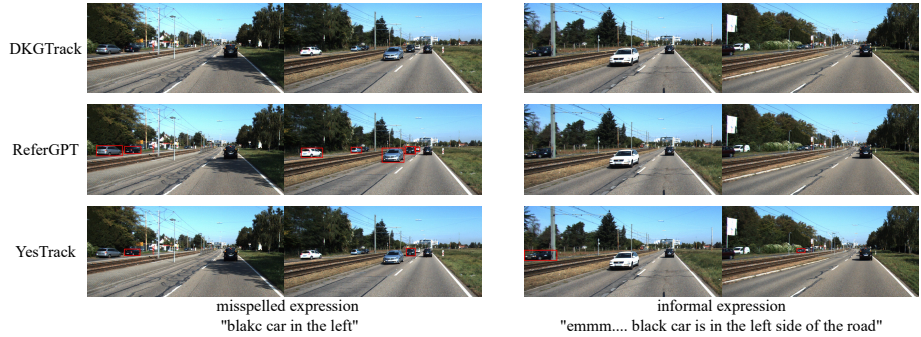}
\caption{Qualitative comparison of our proposed YesTrack against the state-of-the-art DKGTrack and the MLLM-based ReferGPT. The visualization highlights the models' robustness when dealing with noisy natural language inputs, specifically misspelled expressions (e.g., "blakc car in the left") and informal expressions (e.g., "emmm.... black car is in the left side of the road"). YesTrack demonstrates superior capability in comprehending true user intentions despite linguistic noise.}
\label{fig:quai}
\end{figure*}

To further validate the robustness of our proposed method, we conduct a qualitative comparison between YesTrack and two representative baselines: the current state-of-the-art method, DKGTrack, and the MLLM-based approach, ReferGPT. In real-world scenarios, natural language prompts often deviate from standard grammar, frequently containing misspelled expressions or redundant colloquial fillers. As illustrated in Fig.~\ref{fig:quai}, when confronted with such noisy inputs, DKGTrack and ReferGPT often struggle to accurately extract core semantics and easily lose the target. In contrast, YesTrack demonstrates superior anti-interference capabilities, accurately comprehending the user's true intention and stably localizing the target.

\subsection{Limitations}


Despite its efficiency and strong performance, YesTrack is still limited by its two-stage design and the quality of the off-the-shelf tracker. In TRP, referring scores are only evaluated at key frames and inherited by non-key frames, so tracking errors or incorrect MLLM verification at a key frame may propagate until the next re-evaluation. This can cause temporary target loss or distractor tracking. 
Future work will explore a more tightly coupled framework and replace the heuristic TCP with a learnable module to better handle temporal errors and challenging dynamics.

\section{Conclusion}
We propose YesTrack, an efficient RMOT framework that reformulates MLLM inference as binary matching, eliminating text generation latency. With Temporal Confidence Prior and Temporal Reference Propagation, it improves robustness and efficiency in two-stage RMOT and generalizes to MOT (YesTrack-MOT). Experiments show superior performance, demonstrating the effectiveness of discriminative MLLMs for multimodal video understanding. 

\section*{Acknowledgements}

This work was supported by the Natural Science Foundation of China (No. 62372082), the Fundamental Research Funds for the Central Universities (No. ZYGX2024Z017), and Shenzhen Natural Science Foundation (No. JCYJ2024081\\3114206010).
%
%
\bibliographystyle{splncs04}
\bibliography{main}

@String(CVPR  = {IEEE Conf. Comput. Vis. Pattern Recog.})

@String(AAAI  = {AAAI})

@String(ICIP  = {IEEE Int. Conf. Image Process.})

@String(ICASSP=	{ICASSP})

@String(CVPR  = {CVPR})

@String(ICIP  = {ICIP})

@inproceedings{bewley2016simple,
  title={Simple online and realtime tracking},
  author={Bewley, Alex and Ge, Zongyuan and Ott, Lionel and Ramos, Fabio and Upcroft, Ben},
  booktitle={2016 IEEE international conference on image processing (ICIP)},
  pages={3464--3468},
  year={2016},
  organization={Ieee}
}

@inproceedings{wojke2017simple,
  title={Simple online and realtime tracking with a deep association metric},
  author={Wojke, Nicolai and Bewley, Alex and Paulus, Dietrich},
  booktitle={2017 IEEE international conference on image processing (ICIP)},
  pages={3645--3649},
  year={2017},
  organization={IEEE}
}

@inproceedings{zhang2022bytetrack,
  title={Bytetrack: Multi-object tracking by associating every detection box},
  author={Zhang, Yifu and Sun, Peize and Jiang, Yi and Yu, Dongdong and Weng, Fucheng and Yuan, Zehuan and Luo, Ping and Liu, Wenyu and Wang, Xinggang},
  booktitle={European conference on computer vision},
  pages={1--21},
  year={2022},
  organization={Springer}
}

@inproceedings{zeng2022motr,
  title={Motr: End-to-end multiple-object tracking with transformer},
  author={Zeng, Fangao and Dong, Bin and Zhang, Yuang and Wang, Tiancai and Zhang, Xiangyu and Wei, Yichen},
  booktitle={European conference on computer vision},
  pages={659--675},
  year={2022},
  organization={Springer}
}

@article{fang2025associate,
  title={Associate everything detected: Facilitating tracking-by-detection to the unknown},
  author={Fang, Zimeng and Liang, Chao and Zhou, Xue and Zhu, Shuyuan and Li, Xi},
  journal={IEEE Transactions on Image Processing},
  year={2025},
  publisher={IEEE}
}

@inproceedings{wu2023referring,
  title={Referring multi-object tracking},
  author={Wu, Dongming and Han, Wencheng and Wang, Tiancai and Dong, Xingping and Zhang, Xiangyu and Shen, Jianbing},
  booktitle={Proceedings of the IEEE/CVF conference on computer vision and pattern recognition},
  pages={14633--14642},
  year={2023}
}

@article{zhang2024bootstrapping,
  title={Bootstrapping referring multi-object tracking},
  author={Zhang, Yani and Wu, Dongming and Han, Wencheng and Dong, Xingping},
  journal={arXiv preprint arXiv:2406.05039},
  year={2024}
}

@inproceedings{du2024ikun,
  title={ikun: Speak to trackers without retraining},
  author={Du, Yunhao and Lei, Cheng and Zhao, Zhicheng and Su, Fei},
  booktitle={Proceedings of the IEEE/CVF Conference on Computer Vision and Pattern Recognition},
  pages={19135--19144},
  year={2024}
}

@InProceedings{Chamiti_2025_CVPR,
    author    = {Chamiti, Tzoulio and Di Bella, Leandro and Munteanu, Adrian and Deligiannis, Nikos},
    title     = {ReferGPT: Towards Zero-Shot Referring Multi-Object Tracking},
    booktitle = {Proceedings of the IEEE/CVF Conference on Computer Vision and Pattern Recognition (CVPR) Workshops},
    month     = {June},
    year      = {2025},
    pages     = {3888-3897}
}

@inproceedings{li2025language,
  title={Language decoupling with fine-grained knowledge guidance for referring multi-object tracking},
  author={Li, Guangyao and Zhuang, Siping and Jian, Yajun and Yan, Yan and Wang, Hanzi},
  booktitle={Proceedings of the IEEE/CVF International Conference on Computer Vision},
  pages={23626--23635},
  year={2025}
}

@article{Zhao_Hao_Zhang_Liu_Li_Sui_He_Chen_2025, 
title={HFF-Tracker: A Hierarchical Fine-grained Fusion Tracker for Referring Multi-Object Tracking}, 
volume={39}, 
url={https://ojs.aaai.org/index.php/AAAI/article/view/33143}, 
DOI={10.1609/aaai.v39i10.33143}, 
abstractNote={Referring Multi-Object Tracking (RMOT) aims to track multiple objects based on a provided language expression. Although prior studies have sought to accomplish this by integrating an textual module into the multi-object tracker, these methods combine text and image features in a basic way, neglecting the importance of text features. In this study, we propose a Hierarchical Fine-grained text-image Fusion tracker, named HFF-Tracker, which can perform fine-grained fusion of pixel-level visual features and text features across various semantic levels. Specifically, we have devised a Hierarchical Multi-Modal Fusion (HMMF) module to merge text and image features at an early stage in a hierarchical and detailed manner. The Text-Guided Decoder (TGD) is designed to provide the query with prior semantic information during the decoding process. Additionally, we have crafted a Text-Guided Prediction Head (TGPH) that utilizes text information to enhance the performance of the prediction head. Furthermore, we have implemented an adaptive Look-Back training strategy to maximize the utilization of valuable labeled data. Extensive experiments on the Refer-KITTI dataset and the Refer-KITTI-V2 dataset demonstrate that our proposed HFF-Tracker outperforms other state-of-the-art methods with remarkable margins.}, 
number={10}, 
journal={Proceedings of the AAAI Conference on Artificial Intelligence}, 
author={Zhao, Zeyong and Hao, Yanchao and Zhang, Minghao and Liu, Qingbin and Li, Bo and Sui, Dianbo and He, Shizhu and Chen, Xi}, 
year={2025}, 
month={Apr.}, 
pages={10528-10536} }

@article{han2024clip,
  title={CLIP-SCGI: Synthesized caption-guided inversion for person re-identification},
  author={Han, Qianru and He, Xinwei and Liu, Zhi and Liu, Sannyuya and Zhang, Ying and Xiang, Jinhai},
  journal={arXiv preprint arXiv:2410.09382},
  year={2024}
}

@article{wang2024large,
  title={When large vision-language models meet person re-identification},
  author={Wang, Qizao and Li, Bin and Xue, Xiangyang},
  journal={arXiv preprint arXiv:2411.18111},
  year={2024}
}

@article{lu2025llava,
  title={LLaVA-ReID: Selective multi-image questioner for interactive person re-identification},
  author={Lu, Yiding and Yang, Mouxing and Peng, Dezhong and Hu, Peng and Lin, Yijie and Peng, Xi},
  journal={arXiv preprint arXiv:2504.10174},
  year={2025}
}

@article{lin2024echotrack,
  title={Echotrack: Auditory referring multi-object tracking for autonomous driving},
  author={Lin, Jiacheng and Chen, Jiajun and Peng, Kunyu and He, Xuan and Li, Zhiyong and Stiefelhagen, Rainer and Yang, Kailun},
  journal={IEEE Transactions on Intelligent Transportation Systems},
  volume={25},
  number={11},
  pages={18964--18977},
  year={2024},
  publisher={IEEE}
}

@inproceedings{he2024visual,
  title={Visual-linguistic representation learning with deep cross-modality fusion for referring multi-object tracking},
  author={He, Wenyan and Jian, Yajun and Lu, Yang and Wang, Hanzi},
  booktitle={ICASSP 2024-2024 IEEE International Conference on Acoustics, Speech and Signal Processing (ICASSP)},
  pages={6310--6314},
  year={2024},
  organization={IEEE}
}

@article{chen2025multigranularity,
  title={Multigranularity localization transformer with collaborative understanding for referring multiobject tracking},
  author={Chen, Jiajun and Lin, Jiacheng and Zhong, Guojin and Yao, You and Li, Zhiyong},
  journal={IEEE Transactions on Instrumentation and Measurement},
  volume={74},
  pages={1--13},
  year={2025},
  publisher={IEEE}
}

@article{ma2024mls,
  title={Mls-track: Multilevel semantic interaction in rmot},
  author={Ma, Zeliang and Yang, Song and Cui, Zhe and Zhao, Zhicheng and Su, Fei and Liu, Delong and Wang, Jingyu},
  journal={arXiv preprint arXiv:2404.12031},
  year={2024}
}

@article{liang2025cognitive,
  title={Cognitive disentanglement for referring multi-object tracking},
  author={Liang, Shaofeng and Guan, Runwei and Lian, Wangwang and Liu, Daizong and Sun, Xiaolou and Wu, Dongming and Yue, Yutao and Ding, Weiping and Xiong, Hui},
  journal={Information Fusion},
  volume={124},
  pages={103349},
  year={2025},
  publisher={Elsevier}
}

@article{xiao2025temporal,
  title={Temporal-Enhanced Multimodal Transformer for Referring Multi-Object Tracking and Segmentation},
  author={Xiao, Changcheng and Cao, Qiong and Zhong, Yujie and Zhang, Xiang and Wang, Tao and Yang, Canqun and Lan, Long},
  journal={IEEE Transactions on Circuits and Systems for Video Technology},
  year={2025},
  publisher={IEEE}
}

@inproceedings{tran2024mex,
  title={Mex: Memory-efficient approach to referring multi-object tracking},
  author={Tran, Huu-Thien and Pham, Phuoc-Sang and Tran, Thai-Son and Luu, Khoa},
  booktitle={2024 International Conference on Advanced Technologies for Communications (ATC)},
  pages={550--555},
  year={2024},
  organization={IEEE}
}

@misc{bai2025qwen25vltechnicalreport,
      title={Qwen2.5-VL Technical Report}, 
      author={Shuai Bai and Keqin Chen and Xuejing Liu and Jialin Wang 
    and Wenbin Ge and Sibo Song and Kai Dang and Peng Wang and others},
      year={2025},
      eprint={2502.13923},
      archivePrefix={arXiv},
      primaryClass={cs.CV},
      url={https://arxiv.org/abs/2502.13923}, 
}

@misc{bai2025qwen3vltechnicalreport,
      title={Qwen3-VL Technical Report}, 
      author={Shuai Bai and Yuxuan Cai and Ruizhe Chen and Keqin Chen and Xionghui Chen and Zesen Cheng and Lianghao Deng and Wei Ding and Chang Gao and Chunjiang Ge and Wenbin Ge and Zhifang Guo and others},
      year={2025},
      eprint={2511.21631},
      archivePrefix={arXiv},
      primaryClass={cs.CV},
      url={https://arxiv.org/abs/2511.21631}, 
}

@misc{gemmateam2025gemma3technicalreport,
      title={Gemma 3 Technical Report}, 
      author={Gemma Team and Aishwarya Kamath and Johan Ferret and Shreya Pathak and Nino Vieillard and Ramona Merhej and Sarah Perrin and Tatiana Matejovicova and Alexandre Ramé and Morgane Rivière and Louis Rouillard and Thomas Mesnard and Geoffrey Cideron and Jean-bastien Grill and Sabela Ramos and Edouard Yvinec and Michelle Casbon and Etienne Pot and Ivo Penchev and Gaël Liu and Francesco Visin and Kathleen Kenealy and Lucas Beyer and Xiaohai Zhai and Anton Tsitsulin and Robert Busa-Fekete and Alex Feng and Noveen Sachdeva and Benjamin Coleman and others},
      year={2025},
      eprint={2503.19786},
      archivePrefix={arXiv},
      primaryClass={cs.CL},
      url={https://arxiv.org/abs/2503.19786}, 
}

@misc{liu2023visualinstructiontuning,
      title={Visual Instruction Tuning}, 
      author={Haotian Liu and Chunyuan Li and Qingyang Wu and Yong Jae Lee},
      year={2023},
      eprint={2304.08485},
      archivePrefix={arXiv},
      primaryClass={cs.CV},
      url={https://arxiv.org/abs/2304.08485}, 
}

@misc{wang2025internvl35advancingopensourcemultimodal,
      title={InternVL3.5: Advancing Open-Source Multimodal Models in Versatility, Reasoning, and Efficiency}, 
      author={Weiyun Wang and Zhangwei Gao and Lixin Gu and Hengjun Pu and Long Cui and Xingguang Wei and Zhaoyang Liu and Linglin Jing and Shenglong Ye and Jie Shao and Zhaokai Wang and Zhe Chen and Hongjie Zhang and Ganlin Yang and Haomin Wang and Qi Wei and Jinhui Yin and Wenhao Li and Erfei Cui and Guanzhou Chen and Zichen Ding and others},
      year={2025},
      eprint={2508.18265},
      archivePrefix={arXiv},
      primaryClass={cs.CV},
      url={https://arxiv.org/abs/2508.18265}, 
}

@misc{ye2021deeplearningpersonreidentification,
      title={Deep Learning for Person Re-identification: A Survey and Outlook}, 
      author={Mang Ye and Jianbing Shen and Gaojie Lin and Tao Xiang and Ling Shao and Steven C. H. Hoi},
      year={2021},
      eprint={2001.04193},
      archivePrefix={arXiv},
      primaryClass={cs.CV},
      url={https://arxiv.org/abs/2001.04193}, 
}

@misc{hermans2017defensetripletlossperson,
      title={In Defense of the Triplet Loss for Person Re-Identification}, 
      author={Alexander Hermans and Lucas Beyer and Bastian Leibe},
      year={2017},
      eprint={1703.07737},
      archivePrefix={arXiv},
      primaryClass={cs.CV},
      url={https://arxiv.org/abs/1703.07737}, 
}

@misc{luo2019bagtricksstrongbaseline,
      title={Bag of Tricks and A Strong Baseline for Deep Person Re-identification}, 
      author={Hao Luo and Youzhi Gu and Xingyu Liao and Shenqi Lai and Wei Jiang},
      year={2019},
      eprint={1903.07071},
      archivePrefix={arXiv},
      primaryClass={cs.CV},
      url={https://arxiv.org/abs/1903.07071}, 
}

@misc{he2021transreidtransformerbasedobjectreidentification,
      title={TransReID: Transformer-based Object Re-Identification}, 
      author={Shuting He and Hao Luo and Pichao Wang and Fan Wang and Hao Li and Wei Jiang},
      year={2021},
      eprint={2102.04378},
      archivePrefix={arXiv},
      primaryClass={cs.CV},
      url={https://arxiv.org/abs/2102.04378}, 
}

@inproceedings{wang2019rgb,
  title={RGB-infrared cross-modality person re-identification via joint pixel and feature alignment},
  author={Wang, Guan'an and Zhang, Tianzhu and Cheng, Jian and Liu, Si and Yang, Yang and Hou, Zengguang},
  booktitle={Proceedings of the IEEE/CVF international conference on computer vision},
  pages={3623--3632},
  year={2019}
}

@inproceedings{Deruyttere_2019,
   title={Talk2Car: Taking Control of Your Self-Driving Car},
   url={http://dx.doi.org/10.18653/v1/D19-1215},
   DOI={10.18653/v1/d19-1215},
   booktitle={Proceedings of the 2019 Conference on Empirical Methods in Natural Language Processing and the 9th International Joint Conference on Natural Language Processing (EMNLP-IJCNLP)},
   publisher={Association for Computational Linguistics},
   author={Deruyttere, Thierry and Vandenhende, Simon and Grujicic, Dusan and Van Gool, Luc and Moens, Marie-Francine},
   year={2019},
   pages={2088–2098} }

@misc{khoreva2019videoobjectsegmentationlanguage,
      title={Video Object Segmentation with Language Referring Expressions}, 
      author={Anna Khoreva and Anna Rohrbach and Bernt Schiele},
      year={2019},
      eprint={1803.08006},
      archivePrefix={arXiv},
      primaryClass={cs.CV},
      url={https://arxiv.org/abs/1803.08006}, 
}

@misc{nagaraja2016modelingcontextobjectsreferring,
      title={Modeling Context Between Objects for Referring Expression Understanding}, 
      author={Varun K. Nagaraja and Vlad I. Morariu and Larry S. Davis},
      year={2016},
      eprint={1608.00525},
      archivePrefix={arXiv},
      primaryClass={cs.CV},
      url={https://arxiv.org/abs/1608.00525}, 
}

@misc{yu2016modelingcontextreferringexpressions,
      title={Modeling Context in Referring Expressions}, 
      author={Licheng Yu and Patrick Poirson and Shan Yang and Alexander C. Berg and Tamara L. Berg},
      year={2016},
      eprint={1608.00272},
      archivePrefix={arXiv},
      primaryClass={cs.CV},
      url={https://arxiv.org/abs/1608.00272}, 
}

@misc{shinn2023reflexionlanguageagentsverbal,
      title={Reflexion: Language Agents with Verbal Reinforcement Learning}, 
      author={Noah Shinn and Federico Cassano and Edward Berman and Ashwin Gopinath and Karthik Narasimhan and Shunyu Yao},
      year={2023},
      eprint={2303.11366},
      archivePrefix={arXiv},
      primaryClass={cs.AI},
      url={https://arxiv.org/abs/2303.11366}, 
}

@misc{tian2025macgyverlargelanguagemodels,
      title={MacGyver: Are Large Language Models Creative Problem Solvers?}, 
      author={Yufei Tian and Abhilasha Ravichander and Lianhui Qin and Ronan Le Bras and Raja Marjieh and Nanyun Peng and Yejin Choi and Thomas L. Griffiths and Faeze Brahman},
      year={2025},
      eprint={2311.09682},
      archivePrefix={arXiv},
      primaryClass={cs.CL},
      url={https://arxiv.org/abs/2311.09682}, 
}

@misc{sun2024thinkongraphdeepresponsiblereasoning,
      title={Think-on-Graph: Deep and Responsible Reasoning of Large Language Model on Knowledge Graph}, 
      author={Jiashuo Sun and Chengjin Xu and Lumingyuan Tang and Saizhuo Wang and Chen Lin and Yeyun Gong and Lionel M. Ni and Heung-Yeung Shum and Jian Guo},
      year={2024},
      eprint={2307.07697},
      archivePrefix={arXiv},
      primaryClass={cs.CL},
      url={https://arxiv.org/abs/2307.07697}, 
}

@ARTICLE{PC3T,
  author={Wu, Hai and Han, Wenkai and Wen, Chenglu and Li, Xin and Wang, Cheng},
  journal={IEEE Transactions on Intelligent Transportation Systems}, 
  title={3D Multi-Object Tracking in Point Clouds Based on Prediction Confidence-Guided Data Association}, 
  year={2022},
  volume={23},
  number={6},
  pages={5668-5677},
  doi={10.1109/TITS.2021.3055616}}

@article{luiten2021hota,
  title={Hota: A higher order metric for evaluating multi-object tracking},
  author={Luiten, Jonathon and Osep, Aljosa and Dendorfer, Patrick and Torr, Philip and Geiger, Andreas and Leal-Taix{\'e}, Laura and Leibe, Bastian},
  journal={International journal of computer vision},
  volume={129},
  pages={548--578},
  year={2021},
  publisher={Springer}
}

@article{kuhn1955hungarian,
  title={The Hungarian method for the assignment problem},
  author={Kuhn, Harold W},
  journal={Naval research logistics quarterly},
  volume={2},
  number={1-2},
  pages={83--97},
  year={1955},
  publisher={Wiley Online Library}
}

@misc{aharon2022botsortrobustassociationsmultipedestrian,
      title={BoT-SORT: Robust Associations Multi-Pedestrian Tracking}, 
      author={Nir Aharon and Roy Orfaig and Ben-Zion Bobrovsky},
      year={2022},
      eprint={2206.14651},
      archivePrefix={arXiv},
      primaryClass={cs.CV},
      url={https://arxiv.org/abs/2206.14651}, 
}

@InProceedings{Shim_2025_CVPR,
    author    = {Shim, Kyujin and Ko, Kangwook and Yang, Yujin and Kim, Changick},
    title     = {Focusing on Tracks for Online Multi-Object Tracking},
    booktitle = {Proceedings of the IEEE/CVF Conference on Computer Vision and Pattern Recognition (CVPR)},
    month     = {June},
    year      = {2025},
    pages     = {11687-11696}
}

@misc{cao2023observationcentricsortrethinkingsort,
      title={Observation-Centric SORT: Rethinking SORT for Robust Multi-Object Tracking}, 
      author={Jinkun Cao and Jiangmiao Pang and Xinshuo Weng and Rawal Khirodkar and Kris Kitani},
      year={2023},
      eprint={2203.14360},
      archivePrefix={arXiv},
      primaryClass={cs.CV},
      url={https://arxiv.org/abs/2203.14360}, 
}

@misc{robinson2026rfdetrneuralarchitecturesearch,
      title={RF-DETR: Neural Architecture Search for Real-Time Detection Transformers}, 
      author={Isaac Robinson and Peter Robicheaux and Matvei Popov and Deva Ramanan and Neehar Peri},
      year={2026},
      eprint={2511.09554},
      archivePrefix={arXiv},
      primaryClass={cs.CV},
      url={https://arxiv.org/abs/2511.09554}, 
}

@misc{he2020fastreidpytorchtoolboxgeneral,
      title={FastReID: A Pytorch Toolbox for General Instance Re-identification}, 
      author={Lingxiao He and Xingyu Liao and Wu Liu and Xinchen Liu and Peng Cheng and Tao Mei},
      year={2020},
      eprint={2006.02631},
      archivePrefix={arXiv},
      primaryClass={cs.CV},
      url={https://arxiv.org/abs/2006.02631}, 
}
\end{document}